\documentclass[letterpaper]{article} % DO NOT CHANGE THIS
\usepackage[preprint]{aaai2027}  % DO NOT CHANGE THIS
\usepackage[hyphens]{url}  % DO NOT CHANGE THIS
\usepackage{graphicx} % DO NOT CHANGE THIS
\usepackage{natbib}  % DO NOT CHANGE THIS AND DO NOT ADD ANY OPTIONS TO IT
\usepackage{caption} % DO NOT CHANGE THIS AND DO NOT ADD ANY OPTIONS TO IT
\DeclareCaptionStyle{ruled}{labelfont=normalfont,labelsep=colon,strut=off} % DO NOT CHANGE THIS
\graphicspath{{writeup/assets}{assets}}
\providecommand{\inputassets}[1]{%
\IfFileExists{writeup/assets/#1}{\input{writeup/assets/#1}}{\input{assets/#1}}}

\usepackage{subcaption}
\usepackage{amsmath, amssymb}
\usepackage{mathtools}
\let\coloneq\coloneqq
\usepackage{enumitem}
\usepackage{xspace}
\newcommand{\sout}[1]{{\setbox0=\hbox{#1}\rlap{\raisebox{0.35\ht0}{\rule{\wd0}{0.5pt}}}#1}}
\usepackage{tabularray}
\usepackage{dsfont}
\usepackage[capitalise]{cleveref}
\crefname{figure}{Fig.}{Figs.}
\crefname{table}{Table}{Tables}
\crefformat{section}{\S#2#1#3}
\crefname{appendix}{Appendix}{Appendices}

\usepackage{linguex}
\AtBeginDocument{\setlength{\Exlabelsep}{0.5em}}  % gap between example number and text
\usepackage[linguistics]{forest}
\usepackage{tikz-dependency}

\definecolor{cFin}{HTML}{0072B2}    % Wong blue: finite condition
\definecolor{cInf}{HTML}{E69F00}    % Wong orange: infinitival condition
\definecolor{cBare}{HTML}{7F7F7F}   % grey: bare baseline
\newcommand{\textcfin}[1]{\textcolor{cFin}{#1}}
\newcommand{\textcinf}[1]{\textcolor{cInf}{#1}}
\newcommand{\textcbare}[1]{\textcolor{cBare}{#1}}

\newcommand{\bfin}{\beta_{\text{fin}}}
\newcommand{\binf}{\beta_{\text{inf}}}
\newcommand{\bpeak}{\beta^{\text{peak}}}

\newcommand{\dbeta}{\Delta\beta}

\newcommand{\xhead}{\textsuperscript{0\kern-.25ex}}
\newcommand{\xbar}{\ensuremath{'}}

\usepackage[acronym]{glossaries}
\glsdisablehyper

\newacronym{llm}{LLM}{large language model}
\newacronym{ud}{UD}{Universal Dependencies}
\newacronym{mp}{MP}{Minimalist Program}
\newacronym{fdr}{FDR}{false discovery rate}
\newacronym{ci}{CI}{confidence interval}
\newacronym{ecm}{ECM}{exceptional case-marking}
\newacronym{pic}{PIC}{Phase Impenetrability Condition}
\newacronym{ols}{OLS}{ordinary least squares}
\newacronym{udewt}{UD-EWT}{Universal Dependencies English Web Treebank}
\newacronym{uuas}{UUAS}{undirected unlabelled attachment score}

\title{Probing LLMs for Syntactic Structure Beyond Universal Dependencies: A Minimalist Phase Account in English}

\author{
  Yuanhao Chen\quad
  Peter Chin
}
\affiliations{
  Dartmouth College, Hanover, NH 03755, USA \\
  \{yc.th, pc\}@dartmouth.edu
}

\begin{document}
\maketitle

\begin{abstract}
  We show that LLMs encode syntactic distinctions not present in the \gls{ud} tree distances that structural probes are trained to recover.
  On English wh-movement stimuli, we measure the probe distance between an embedded subject and its verb, whose \gls{ud} tree distance is invariant across conditions.
  That distance is \emph{shorter} than baseline when the embedded clause is finite and \emph{longer} when it is infinitival --- a within-clause \emph{sign asymmetry} present in all 13 models across four families we test, at a majority of layers.
  No account based on \gls{ud} distance, linear order, or monotone structural complexity can produce a sign reversal, while Minimalist phase theory can.
  Holding the matrix verb fixed while varying only the complement type reproduces the same finite--infinitival ordering in every model, ruling out a lexical-semantic explanation.
  The cross-clause pair separately reproduces the phase-count ordering of earlier work, validating the probes.
  A single activation patch, interchanging the clause-selecting matrix verb, moves the two pairs in \emph{opposite} directions in 8 of 13 models.
  Because the causal intervention leaves the target string unchanged, it also rules out the clause-length and surface-cue explanations.
  Together these results reveal syntactic structure in LLMs beyond the \gls{ud} probe target, and a Minimalist phase account is consistent with it.
\end{abstract}

% Reset so first body occurrence re-expands (abstract is standalone).
\glsresetall

\section{Introduction}
\label{sec:intro}

Structural probing has established that \glspl{llm} encode syntactic structure in their hidden representations \citep{hewittStructuralProbeFinding2019,manningEmergentLinguisticStructure2020}.
These probes train on \gls{ud} tree distances as a consistent, broadly applicable annotation standard \citep{demarneffeUniversalDependencies2021}, but explicitly not a generative grammar.
Neural language models have become a testbed for formal linguistic theory, from early recurrent models \citep{marvinTargetedSyntacticEvaluation2018,wilcoxWhatRNNLanguage2018} to transformers \citep{kennedyEvidenceGenerativeSyntax2025,kennedyEvidenceHierarchicallyComplexSyntactic2025}.
This raises a question that \gls{ud}-like parsing cannot answer by construction: Do \glspl{llm} encode formal-syntactic abstractions (phase boundaries, phase-internal cohesion) that \gls{ud} is not designed to represent?

We evaluate structural probes on English wh-movement stimuli constructed and verified per item so that \gls{ud}-tree distances between the probed word pairs are invariant across conditions (\cref{sec:stimuli}).
The three conditions --- a bare small clause (an embedded subject and bare verb, with no embedded tense or complementiser), an infinitival, and a finite complement --- are ordered by the number of \gls{mp} phase boundaries the wh-element crosses.
The design follows \citet{kennedyEvidenceGenerativeSyntax2025,kennedyEvidenceHierarchicallyComplexSyntactic2025}, who use \gls{ud}-invariant complement-size contrasts to probe hierarchical structure: With \gls{ud} distance held fixed, any condition effect must reflect structure beyond the \gls{ud} distance.

Our central result is a within-clause \emph{sign asymmetry}.
The structural-probe distance between an embedded subject and the embedded verb reverses sign across conditions: smaller than the bare baseline in the finite condition, larger in the infinitival condition.
No surface property predicts this.
\Gls{ud} distance between the two words is invariant across conditions; linear word distance predicts no finite effect (the two tokens are adjacent in both bare and finite); and a monotone structural-complexity account predicts larger distances in both non-baseline conditions.
The pattern is instead predicted by phase-internal cohesion --- an \gls{mp} abstraction invisible to the \gls{ud} probe target by construction (\cref{sec:esubj-evb}).
We make three contributions.
\begin{enumerate}[topsep=1pt,itemsep=0pt,leftmargin=*]

  \item \textbf{A within-clause sign asymmetry beyond \gls{ud} (\cref{sec:esubj-evb}).}
    On a probe pair whose \gls{ud} distance is invariant across conditions, we find $\bfin < 0$ and $\binf > 0$ (writing $\beta$ for a condition's change in probe distance from the bare baseline), each at a majority of layers, in all 13 \glspl{llm} from four families.
    A same-verb control, holding the matrix verb fixed while varying only complement type, reproduces this ordering in all 13 models and rules out a lexical-semantic explanation.
    The reversal is predicted by \gls{mp} phase-internal cohesion and is invisible to the \gls{ud} tree-distance target by construction.

  \item \textbf{Opposite-signed causal effects (\cref{sec:causal}).}
    A single activation patch, interchanging the clause-selecting matrix verb, shifts the cross-clause and within-clause probe pairs in \emph{opposite} directions.
    Because the causal intervention leaves the target string unchanged, it removes the confounds an observational contrast cannot: It shows the within-clause asymmetry tracks phase status, not clause length, subject--verb agreement, or the intervening \textit{to}.

  \item \textbf{A validated cross-clause ordering across a broad panel (\cref{sec:observational}).}
    We reproduce and extend the cross-clause phase-count ordering of \citet{kennedyEvidenceHierarchicallyComplexSyntactic2025} --- the finite complement yields the largest \textsc{wh-esubj} separation, with $\bfin > 0$ and $\bfin > \binf$ in all 13 models.
    Across four families and 1--27B parameters --- a broader panel than the encoder-only setup of prior structural-probe work on generative syntax --- both this ordering and the sign asymmetry hold, indicating the effects are not model-specific.
    The ordering also validates that our probes detect phase boundaries and supplies one of the two responses in the causal opposite-signed effect.

\end{enumerate}

\section{Background and Related Work}
\label{sec:background}

\paragraph{Phase theory.}
\Cref{app:syntax-glossary} glosses syntactic terms used throughout.
We treat clause structure in the framework of phase theory \citep{chomskyMinimalistInquiriesFramework2000,chomskyDerivationPhase2001}.
A \emph{phase} is a syntactic domain whose complement domain becomes inaccessible to higher derivational operations once the phase is complete.
The two phase heads in English are \textit{v}\xhead, the light-verb head projecting a \textit{v}P, and C, the head of CP.
The wh-element crosses the matrix \textit{v}P in every condition; a finite embedded clause adds two further phases along its path (the embedded \textit{v}P and CP), an infinitival adds one (the embedded \textit{v}P), and a bare small-clause complement adds none.
Its movement path therefore spans three phase boundaries with a finite complement, two with an infinitival, and one with a bare complement.

We adopt the traditional analysis on which a bare perception or causative complement is a small clause projecting only a lexical VP and its subject \citep{stowellOriginsPhraseStructure1981}, and so introduces no embedded phase; whether such complements instead project an eventive \textit{v}P (hence a phase) is debated \citep{felserVerbalComplementClauses1999}, but the finite-versus-infinitival ordering that drives our central asymmetry (finite tighter than infinitival on \textsc{esubj-evb}) compares the two non-bare conditions and is unaffected by the bare baseline's exact phase status.

The \gls{pic} restricts cross-phase operations to material at the phase edge \citep{chomskyDerivationPhase2001}, forcing successive-cyclic wh-movement to transit each phase edge \citep{urkSuccessiveCyclicitySyntax2020}.
Two predictions from this framework motivate our experimental design.
The number of phase boundaries between the wh-position (matrix Spec,CP, after movement) and the base-position wh-copy at the embedded verb's complement tracks the complement type: bare $<$ infinitival $<$ finite (\cref{sec:stimuli}).
Arguments inside the same phase are structurally more cohesive than arguments separated by a phase boundary, even when their dependency-tree distance is identical (\cref{sec:esubj-evb}).
The first motivates the cross-clause \textsc{wh-esubj} probe pair; the second motivates the within-clause \textsc{esubj-evb} pair (\cref{sec:stimuli}).

\paragraph{Phase-internal cohesion.}
Beyond restricting cross-phase operations, phases are standardly taken to be the units of cyclic spell-out \citep{uriagerekaMultipleSpellOut1999,foxCyclicLinearizationSyntactic2005} and the domains within which agreement, case, and binding are computed \citep{boskovicLocalityMotivationMove2007,lee-schoenfeldBindingPhasesLocality2008}; sentence-processing work likewise finds integrative computation consolidating within-clause material at clause boundaries \citep{raynerEffectClauseWrapUp2000}.
None of this is a representational claim about \glspl{llm}, but each gives a converging reason to expect items inside one completed phase to be represented more cohesively than items split across a phase boundary --- the hypothesis we call \emph{phase-internal cohesion} and test in \cref{sec:esubj-evb}.

\paragraph{Structural probes and prior evaluations.}
A structural probe is a linear map from a model's contextualised hidden states to a Euclidean space, trained so that pairwise distances approximate gold tree distances between words \citep{hewittStructuralProbeFinding2019,manningEmergentLinguisticStructure2020}; because training and evaluation both operate in \gls{ud}, structural probing is by construction sensitive only to \gls{ud}-grounded distinctions.
Because an expressive probe can achieve high accuracy by fitting the mapping itself rather than reading structure off the representation, probe selectivity is a standard concern \citep{hewittDesigningInterpretingProbes2019}; our design addresses it by holding the \gls{ud} target invariant across conditions, so any condition effect must originate in the representation rather than in the probe's \gls{ud}-fitting capacity.

\Citet{kennedyEvidenceGenerativeSyntax2025,kennedyEvidenceHierarchicallyComplexSyntactic2025} are the closest precedent: They evaluate structural probes on generative-syntax contrasts (subject raising vs.\ control; complement size) chosen so that \gls{ud} does not directly encode the distinction.
We share their design of holding \gls{ud} distance invariant across conditions (verified per item, \cref{sec:stimuli}), but our contribution differs in kind: a within-clause \emph{sign asymmetry}, a phase-internal-cohesion account of it, and a \emph{causal} test.
Their evidence also rests on a few small models (BERT, RoBERTa, and GPT-2, plus two sub-2B Qwen-2.5 models), whereas we test the phenomenon on 13 decoder-only \glspl{llm} of 1--27B parameters from four contemporary families --- the autoregressive generation now in widespread use.

A parallel behavioural tradition assesses syntactic knowledge from model predictions, via minimal-pair acceptability \citep{marvinTargetedSyntacticEvaluation2018} or surprisal on controlled items \citep{wilcoxWhatRNNLanguage2018,kooSuccessivecyclicMovementHumans2026}; \citet{agarwalMechanismsVsOutcomes2025} show across 32 models that structural-probe accuracy does not predict minimal-pair performance, indicating that representational and behavioural methods recover different aspects of syntactic knowledge.

Whether a probed representation is causally \emph{active} --- whether intervening on it changes downstream representations, in the causal-abstraction sense \citep{geigerCausalAbstractionTheoretical2023} --- is distinct from whether the model's \emph{behaviour} relies on it; we test the former by activation patching (replacing an internal representation with one from a counterfactual input and measuring the downstream change, \cref{sec:causal}) and leave the latter open.

\paragraph{Effect size.}
Our probes are trained on the \gls{udewt} \citep{silveiraGoldStandardDependency2014} with undirected dependency-tree distance as the gold target, then evaluated on our wh-movement stimuli without further fitting.
We report an effect size $\beta$: the estimated condition-vs.-bare difference in probe distance for a designated word pair (\cref{sec:probing-setup}).
Because our stimuli are constructed so that \gls{ud}-tree distance between probed word pairs is invariant across conditions, verified per item (\cref{sec:stimuli}), a non-zero $\beta$ cannot be a re-encoding of the probe's training target; it must reflect structural information that the \gls{ud} tree distance does not encode.

\section[Stimuli and Methods]{Stimuli and Methods\footnote{Code and stimuli: \url{https://anonymous.4open.science/r/syntax-probe-147D}.}}
\label{sec:methods}

\subsection{Stimuli}
\label{sec:stimuli}

Each item is a triple of wh-questions sharing the wh-element (\textit{what}), the embedded-subject, and the embedded-verb, with the matrix verb and complement type varying by condition:

\ex.
\a.[\bf Bare: ] What did she see him eat?
\b.[\bf Infinitival: ] What did she expect him to eat?
\c.[\bf Finite: ] What did she think he ate?

\begin{figure}[t]
  \captionsetup[subfigure]{justification=raggedleft, singlelinecheck=false}
  \begin{subfigure}[t]{0.32\linewidth}
    \centering
    \scalebox{0.9}{%
      \begin{forest}
        for tree={s sep=4pt, inner sep=0.6pt, l=14pt, l sep=4pt, font=\footnotesize}
        [CP
          [what$_i$]
          [C\xbar
            [did]
            [TP
              [she]
              [\textbf{\textit{v}P}
                [see]
                [VP, for tree={text=cBare}
                  [him]
                  [V\xbar
                    [eat]
                    [\sout{what$_i$}]
                  ]
                ]
              ]
            ]
          ]
        ]
    \end{forest}}%
    \makebox[0pt][l]{\vphantom{\scalebox{0.9}{%
          \begin{forest}
            for tree={s sep=4pt, inner sep=0.6pt, l=14pt, l sep=4pt, font=\footnotesize}
            [CP
              [what$_i$]
              [C\xbar
                [did]
                [TP
                  [she]
                  [\textit{v}P
                    [think]
                    [CP
                      [$\emptyset$]
                      [TP
                        [he]
                        [\textit{v}P
                          [ate]
                          [\sout{what$_i$}]
                        ]
                      ]
                    ]
                  ]
                ]
              ]
            ]
        \end{forest}}%
    }}%
    \caption{Bare}
  \end{subfigure}\kern-1ex
  \begin{subfigure}[t]{0.335\linewidth}
    \centering
    \scalebox{0.9}{%
      \begin{forest}
        for tree={s sep=4pt, inner sep=0.6pt, l=14pt, l sep=4pt, font=\footnotesize}
        [CP
          [what$_i$]
          [C\xbar
            [did]
            [TP
              [she]
              [\textbf{\textit{v}P}
                [expect]
                [TP, for tree={text=cInf}
                  [him]
                  [T\xbar
                    [to]
                    [\textbf{\textit{v}P}
                      [eat]
                      [\sout{what$_i$}]
                    ]
                  ]
                ]
              ]
            ]
          ]
        ]
    \end{forest}}
    \caption{Infinitival}
  \end{subfigure}\kern-1ex
  \begin{subfigure}[t]{0.335\linewidth}
    \centering
    \scalebox{0.9}{%
      \begin{forest}
        for tree={s sep=4pt, inner sep=0.6pt, l=14pt, l sep=4pt, font=\footnotesize}
        [CP
          [what$_i$]
          [C\xbar
            [did]
            [TP
              [she]
              [\textbf{\textit{v}P}
                [think]
                [\textbf{CP}, for tree={text=cFin}
                  [$\emptyset$]
                  [{TP}
                    [he]
                    [\textbf{\textit{v}P}
                      [ate]
                      [\sout{what$_i$}]
                    ]
                  ]
                ]
              ]
            ]
          ]
        ]
    \end{forest}}
    \caption{Finite}
  \end{subfigure}

  \begin{center}\footnotesize
    \begin{tblr}{
        colspec = {l ccc},
        row{1}  = {font=\bfseries},
        hline{1,Z} = {0.8pt},
        hline{2}   = {0.4pt},
        rowsep  = 1.2pt,
        colsep  = 5pt,
      }
      & Bare & Infinitival & Finite \\
      Phase boundaries (\textsc{wh} path) & $1$ & $2$ & $3$ \\
      \gls{ud} distance: \textsc{wh-esubj}  & $2$ & $2$ & $2$ \\
      \gls{ud} distance: \textsc{esubj-evb} & $1$ & $1$ & $1$ \\
      Prediction: $\beta$ \textsc{wh-esubj}  & / & $+$ & $+$ \\
      Prediction: $\beta$ \textsc{esubj-evb} & / & $+$ & $-$ \\
    \end{tblr}
  \end{center}

  \caption{
    Schematic phrase structures of the three conditions and what they predict for the two probe pairs.
    The embedded clause is colour-coded; phases crossed by the wh-element are in \textbf{bold}.
    The lower wh-copy is shown struck through at its base position, and intermediate bar levels and shared matrix projections are omitted.
    The table reads off, per condition, the phase boundaries on the wh-movement path, the within-item \gls{ud}-tree distance for each probe pair, and the predicted sign of the condition-vs.-bare change in probe distance.
    Both \gls{ud} distances are invariant across conditions ($2$ edges for \textsc{wh-esubj}, $1$ for \textsc{esubj-evb}, in the intended parse), so any non-zero $\beta$ reflects structure beyond the \gls{ud} distance.
    The two pairs diverge --- \textsc{wh-esubj} grows with phase count ($\bfin > \binf > 0$), whereas \textsc{esubj-evb} reverses sign ($\bfin < 0$, $\binf > 0$).
  }
  \label{fig:trees}
\end{figure}

The matrix verb is drawn from a condition-specific class.
The \emph{bare} matrix verb is a perception or causative verb (\textit{see}, \textit{watch}, \textit{make}, \textit{let}) selecting a bare small-clause complement \citep{stowellOriginsPhraseStructure1981}.
The \emph{infinitival} matrix is an \gls{ecm} or object-control verb (\textit{expect}, \textit{want}, \textit{allow}, \textit{need}) selecting an infinitival TP complement \citep{chomskyBarriers1986}.\footnote{%
We use \emph{embedded subject} for the accusative pronoun that is the logical subject of the embedded verb.
Syntactically it is the embedded subject under \gls{ecm} (\emph{expect}) and the matrix object controlling a null PRO subject under object control (\emph{want}, \emph{allow}, \emph{need}); the phase-cohesion mechanism of \cref{sec:esubj-evb} is stated for the \gls{ecm} case.
The two subclasses yield comparable $\binf$, positive in all 13 models (\cref{app:ecm-control}), so we treat the infinitival as a single class.
On a raising-to-object analysis the pronoun attaches to the matrix verb, enlarging the within-clause distance; the design relies only on that distance being invariant across conditions, which we verify per item, not on its absolute value.%
}
The \emph{finite} matrix is a bridge verb (\textit{think}, \textit{believe}, \textit{claim}, \textit{say}, \textit{know}, \textit{suppose}, \textit{report}) selecting a finite CP complement with a null complementiser \citep{erteschik-shirNatureIslandConstraints1973}.
In the bare and infinitival conditions the embedded subject bears accusative case and the embedded verb appears in its base form; in the finite condition the embedded subject bears nominative case and the embedded verb appears in the simple past.

The three conditions are ordered by the number of phase boundaries crossed along the wh-movement path \citep[\cref{fig:trees};][]{chomskyMinimalistInquiriesFramework2000,chomskyDerivationPhase2001}: bare small clause $<$ infinitival TP $<$ finite CP.
We treat bare as the structural baseline: A contrast against bare measures the representational consequences of the additional clause structure.

We tag three positions per stimulus (\emph{what}, embedded subject, embedded verb) and measure two probe distances: \textsc{wh-esubj} (\emph{what} to embedded subject), capturing cross-clause structural depth as a function of complement type; and \textsc{esubj-evb} (embedded subject to embedded verb), capturing within-clause cohesion.

\paragraph{UD-distance invariance.}
For probe-distance differences to be interpretable, the \gls{ud}-tree distance between each probe pair must be constant across conditions within each item; otherwise differences in $\beta$ would be confounded with \gls{ud}-level differences.
We construct the stimuli from templates whose intended \gls{ud} parse gives \textsc{esubj-evb} a single \texttt{nsubj} edge and \textsc{wh-esubj} a two-edge path, and we verify per item that the automatic parser keeps each distance constant across conditions.
Parsing all $3{,}000$ stimuli with spaCy's \texttt{en\_core\_web\_trf} model \citep[RoBERTa-based;][]{Honnibal_spaCy_Industrial-strength_Natural_2020} confirms the design --- \textsc{esubj-evb} is fully invariant for $98.8\%$ of items ($99.7\%$ in the same-verb control), and no item exceeds a one-edge spread.
The residual one-edge spread falls on the wh-pairs, where the parser assigns a \emph{shorter} infinitival path than finite; because this shortens rather than lengthens the infinitival distance, it runs against the predicted direction of the cross-clause effect and cannot inflate the false-positive rate.
The check is scheme-independent --- spaCy's native Stanford-style scheme coincides with \gls{ud} on the \texttt{nsubj}, direct-object, and clausal-complement edges that fix our pair distances, differing only in constructions absent from our stimuli (\cref{sec:stimulus-appendix} gives per-pair rates, the parse figure, and the lexicon).
Since our prediction is that probe distance \emph{increases} with structural depth, this directional mismatch cannot manufacture the predicted effect.

\paragraph{Worked example: what each condition predicts.}
The two probe pairs yield two independent, phase-based predictions across the three conditions (\cref{fig:trees}).
For the \textsc{wh-esubj} pair, the wh-element transits every phase boundary on its movement path, so more boundaries mean greater representational separation: bare ($1$) $<$ infinitival ($2$) $<$ finite ($3$), predicting $\bfin > \binf > 0$.
For the \textsc{esubj-evb} pair, they are separated by a phase edge in both the infinitival and finite conditions, loosening it ($\binf > 0$).
But only in the finite condition, they share an extra phase (lower CP), and phase cohesion compensates in the other direction ($\bfin < 0$).
The two pairs therefore carry \emph{opposite} finite-condition signs, the contrast the causal experiment later exploits (\cref{sec:causal}).

\paragraph{Lexicon and item generation.}
The combinatorial lexicon comprises $7$ matrix subjects, $7$ embedded subjects (in matched accusative/nominative pairs), $4$ bare-class matrix verbs, $4$ infinitival-class verbs, $7$ bridge verbs, and $20$ embedded transitive verbs with inanimate-compatible objects.
Items are constrained so that the matrix and embedded subjects differ and so that the three matrix verbs are mutually distinct within an item.
The combination yields on the order of $10^5$ candidate items; we use a fixed seed-controlled sample of $1{,}000$ items, yielding $3{,}000$ stimuli (one per condition per item).

\subsection{Probing Setup}
\label{sec:probing-setup}

\paragraph{Probe training.}
We follow the structural-probe protocol of \citet{hewittStructuralProbeFinding2019}.
For each transformer layer $\ell$, we obtain a per-word representation $h_w^{(\ell)} \in \mathbb{R}^d$ by mean-pooling the subword-token hidden states at layer $\ell$ for each word $w$ ($d$ is the model's hidden size).
We then train a linear projection $B^{(\ell)} \in \mathbb{R}^{r \times d}$ with $r = 64$, defining the probe distance between any two words $u, v$ at layer $\ell$ as
\[
  d_B^{(\ell)}(u, v) \;\coloneq\; \big\lVert B^{(\ell)} \big(h_u^{(\ell)} - h_v^{(\ell)}\big) \big\rVert_2^2,
\]
the squared L2 norm of the projected difference \citep[eq.\,1]{hewittStructuralProbeFinding2019}.
$B^{(\ell)}$ is fit by minimising L1 loss between $d_B^{(\ell)}(u, v)$ and the gold undirected dependency-tree distance between $u, v$ on the \gls{udewt} training corpus \citep{silveiraGoldStandardDependency2014}, using Adam (learning rate $10^{-3}$, batch size $256$) for up to $100$ epochs with learning-rate decay on plateau (factor $0.1$, patience $1$, up to $4$ resets).
Input activations are standardised per training corpus before projection.
The resulting per-layer probes are then evaluated on our wh-movement stimuli without further fitting.

\paragraph{Effect-size estimation.}
For each (model, layer~$\ell$, pair) we fit one treatment-coded \gls{ols} regression
\begin{align*}
  d^{(\ell)}_{i,k} \;=\;\; & \beta_0^{(\ell)} + \bfin^{(\ell)} \cdot \mathds{1}\{c_{i,k} = \text{fin}\} \\
  & \phantom{\beta_0^{(\ell)}} + \binf^{(\ell)} \cdot \mathds{1}\{c_{i,k} = \text{inf}\} + \varepsilon^{(\ell)}_{i,k},
\end{align*}
where $d^{(\ell)}_{i,k}$ is the probe distance $d_B^{(\ell)}$ for the relevant word pair in stimulus $k$ of item $i$, $c_{i,k}$ is the condition, $\mathds{1}\{\cdot\}$ is the indicator function, $\varepsilon^{(\ell)}_{i,k}$ is the error term, and bare is the reference category (so $\beta_0^{(\ell)}$ is the bare-condition mean probe distance).
Empirically $\beta_0$ is on the order of 2--3 across models and layers, so the condition effects $\beta$ we report (roughly $0.1$--$0.5$) are modest fractions of that baseline (about 5--20\%).
The primary coefficients are the contrast slopes $\bfin^{(\ell)}$ (finite\,$-$\,bare) and $\binf^{(\ell)}$ (infinitival\,$-$\,bare).
The three conditions of an item share lexical content, so we fit the regression with cluster-robust (CR1) standard errors clustered on the item ($1{,}000$ clusters), rather than treating the $3{,}000$ stimuli as independent.
We apply Benjamini--Hochberg \gls{fdr} correction \citep{benjaminiControllingFalseDiscovery1995} across all (layer\,$\times$\,pair\,$\times$\,contrast) tests within each model.

\subsection{Layer Reporting}
\label{sec:reference-layer}

Structural-probing effects vary across layers, and reporting only a per-model peak $\bpeak = \max_\ell \beta^{(\ell)}$ \citep{hewittStructuralProbeFinding2019,manningEmergentLinguisticStructure2020} is opportunistic, especially when different contrasts peak at different layers (in our panel, $0$ of $13$ models share a peak layer for $\bfin$ and $\binf$ on \textsc{wh-esubj}).
We therefore report each effect \emph{breadth-first}: the fraction of layers at which it holds in the predicted direction under \gls{fdr} correction, a summary that privileges no single layer.

When a per-pair anchor is needed to site the causal intervention's measurement (\cref{sec:causal}), we use the \emph{reference layer}: $L^\ast \coloneq \operatorname*{\arg\max}_\ell \bfin^{(\ell)}$ for the cross-clause pair, and $\operatorname*{\arg\max}_\ell (\binf^{(\ell)} - \bfin^{(\ell)})$ for the within-clause pair --- in each case the layer at which that pair's effect is strongest.
The causal intervention reads $\dbeta$ (the change in a pair's contrast induced by activation patching) within the \emph{reference band} $L^\ast \pm 1$ (the reference layer and its immediate neighbours, floored at layer~3 so the patch site stays strictly upstream), with $L^\ast$ fixed by the observational effect $\beta$, not by the causal effect $\dbeta$.

The two effects call for different summaries.
The observational effect is broad across layers, so we report the layer fraction above.
The causal effect is instead layer-localised (concentrating near $L^\ast$), so we report its \emph{dominant} in-band value (the largest-magnitude significant $\dbeta$ in the band) rather than a layer fraction; at $n = 1{,}000$ a layer fraction would let a tiny opposite-signed blip one layer away, significant only at this sample size, weigh equally with the main effect.
Significance of each in-band $\dbeta$ comes from a clustered bootstrap over items ($n = 1{,}000$ resamples, percentile $95\%$ \gls{ci}), with a $\dbeta$ significant when its \gls{ci} excludes $0$.

\subsection{Models}
\label{sec:models}

We evaluate on 13 decoder-only language models from four families: Gemma-3 \citep{teamGemma3Technical2025}, Llama-3 \citep{grattafioriLlama3Herd2024}, Mistral \citep{jiangMistral7B2023}, and Qwen \citep{qwenQwen25TechnicalReport2025,yangQwen3TechnicalReport2025}.
All models are base (non-instruction-tuned) pretrained checkpoints, accessed via the Hugging\-Face Transformers library; the full panel is in \cref{app:models}.
Probe training and evaluation were run on a single NVIDIA RTX~PRO~6000 Blackwell GPU; total wall-clock time was under two hours.

\section{Experiments}
\label{sec:experiments}

\subsection{Observational Validation: The Cross-Clause Phase-Count Ordering}
\label{sec:observational}

\begin{figure}[t]
  \centering
  \includegraphics[width=\linewidth]{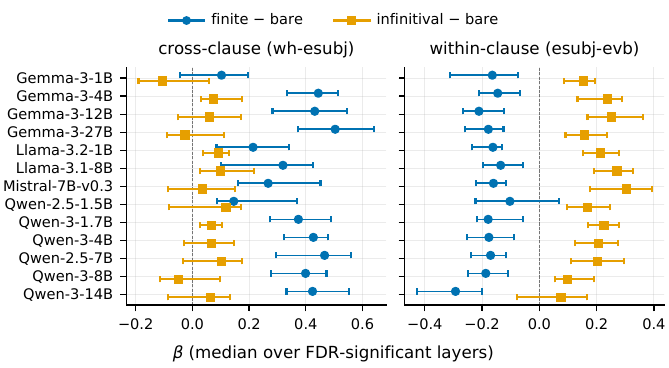}
  \caption{
    Observational phase-structure effects across all 13 models, on both probe pairs.
    Left: the cross-clause \textsc{wh-esubj} pair; right: the within-clause \textsc{esubj-evb} pair.
    For each model and contrast (finite\,$-$\,bare, infinitival\,$-$\,bare) the marker is the median $\beta$ across the layers \gls{fdr}-significant for that model, pair, and contrast, and the whisker spans their interquartile range; a hollow marker at $\beta = 0$ denotes no \gls{fdr}-significant layer.
    The cross-clause pair shows the phase-count ordering ($\bfin > \binf$, with $\bfin > 0$ in every model); the within-clause pair shows the sign asymmetry ($\bfin < 0$, $\binf > 0$) in every model.
  }
  \label{fig:phase-forest}
\end{figure}

We first confirm that our probes detect phase boundaries at all, on the cross-clause \textsc{wh-esubj} pair, replicating and extending \citet{kennedyEvidenceHierarchicallyComplexSyntactic2025} across the panel.
Estimating $\bfin^{(\ell)}$ and $\binf^{(\ell)}$ at every layer and summarising each contrast breadth-first (\cref{sec:reference-layer}), we find the finite condition yields the largest separation --- $\bfin > 0$ and $\bfin > \binf$ in all 13 models (\cref{fig:phase-forest}, left) --- so each added phase boundary between the wh-filler and the embedded subject increases their representational separation, most strongly for the finite complement.
The finite--bare contrast is broad and reliable.
The panel median of \gls{fdr}-significant positive layers for $\bfin$ is $94\%$ (five of 13 models reach $\geq 97\%$; lowest $63\%$, Gemma-3-1B), with the median $\bfin$ across \gls{fdr}-significant layers ranging from $+0.10$ (Gemma-3-1B) to $+0.50$ (Gemma-3-27B; \cref{fig:phase-forest}, left).
The smaller one-phase infinitival step $\binf$ is weaker and more layer-localised: positive at a majority of \gls{fdr}-significant layers in 10 of 13 models, with a panel-median marker near zero or slightly negative in the other three (Gemma-3-1B, Gemma-3-27B, and Qwen-3-8B; \cref{fig:phase-forest}, left).
This is expected, because the infinitival is the one condition whose \textsc{wh-esubj} \gls{ud} distance is not strictly invariant.
The parser assigns a shorter path in a minority of items, working against $\binf$, whereas the finite--bare invariance (both $2$ edges) is stable and isolates a clean beyond-\gls{ud} effect.
The basis for $\binf > 0$ is derivational: The phrase-structure distance from \emph{what} to the embedded subject is equal in the bare and infinitival conditions, but the infinitival subject is base-merged at the phase edge (Spec,\textit{v}P) while the bare subject is not (\cref{app:full-trees}; \citealt{chomskyDerivationPhase2001}).
This ordering is the observational counterpart of one of the two responses in the causal opposite-signed effect (\cref{sec:causal}); the novel result is the within-clause pair, which we focus on next.

\subsection{Within-Clause Cohesion: The Sign Asymmetry}
\label{sec:esubj-evb}

We turn from the cross-clause pair to the within-clause pair --- the embedded subject and the embedded verb (\textsc{esubj-evb}) --- whose \gls{ud}-tree distance is invariant across conditions, a single \texttt{nsubj} edge in the intended parse (\cref{fig:trees}).
Here we find a robust \emph{sign asymmetry}: $\bfin < 0 < \binf$.
Reported breadth-first (\cref{sec:reference-layer}), every one of the 13 models, across all four families, shows $\bfin < 0$ and $\binf > 0$ at a majority of layers (per-model values in \cref{sec:obs-forest}).
Across the panel the median $\beta$ over \gls{fdr}-significant layers is negative for $\bfin$ and positive for $\binf$ in every model, with magnitudes of roughly $0.1$--$0.3$ on each side (\cref{fig:phase-forest}, right).
Because the two contrasts have opposite predicted signs, a shared reference layer does not apply; the asymmetry is a property of the layer profile, not of any single layer.

\paragraph{Three observations rule out simpler accounts.}
First, the \gls{ud} distance between the two words is invariant across conditions (\cref{fig:trees}), so a pure \gls{ud}-decoding probe would yield $\beta \approx 0$.
Second, the linear distance between the two words is $1$ in bare and finite but $2$ in infinitival (the \emph{to} marker intervenes).
A surface-linear-distance heuristic predicts $\binf > 0$ and $\bfin \approx 0$; the observed \emph{negative} $\bfin$ cannot come from linear distance.
Third, a monotone structural-complexity heuristic (more tree structure between the two words means larger probe distance) predicts $\binf \gtrsim \bfin \gtrsim 0$ (inequality or equality depending on simplified structure, \cref{fig:trees}, or full structure, \cref{app:full-trees}), with either interpretation yielding $\bfin \gtrsim 0$.
The observed \emph{negative} $\bfin$ directly contradicts these predictions.

One alternative these observations do \emph{not} address is subject--verb agreement --- the finite condition alone pairs a nominative subject with a tensed verb, so $\bfin < 0$ could reflect an agreement relation rather than phase-internal cohesion.
The observational contrast cannot remove this confound; the causal test in \cref{sec:causal} is designed to exclude it.

\paragraph{A same-verb control.}
A further alternative is that the asymmetry reflects the lexical semantics of the matrix verb rather than clause structure: Our finite conditions use attitude verbs and our infinitival conditions expectation and desire verbs.
To rule this out we ran a same-verb control that fixes the matrix verb at \textit{expect}, which independently licenses both a finite complement (\emph{What did she expect he ate?}) and an infinitival one (\emph{What did she expect him to eat?}).
It is the rare verb admitting both with matched aspect; epistemic alternatives such as \textit{believe} require a perfect infinitival (``believe him to \emph{have} eaten,'' not ``believe him to eat''), which would reintroduce an aspectual confound.
A same-verb design admits no bare condition, since \textit{expect} takes no bare complement, so only the finite--infinitival difference is strictly verb-matched; we therefore test the verb-matched contrast $\binf - \bfin$, which cancels the bare baseline.
Holding the verb fixed, the infinitival condition yields a larger \textsc{esubj-evb} probe distance than the finite condition at a majority of layers in all 13 models (\gls{fdr}-corrected; median $\binf - \bfin$ between $0.19$ and $0.39$), reproducing the ordering above with matrix-verb identity and lexical semantics held constant (\cref{sec:b1-appendix}).

\paragraph{From the PIC to representations.}
The \gls{pic} \citep{chomskyMinimalistInquiriesFramework2000,chomskyDerivationPhase2001} is a constraint on syntactic derivation.
It determines which operations the grammar can perform across phase boundaries, and is not a claim about representational geometry in itself.
To connect phase theory to the observed sign asymmetry we therefore need an additional assumption.
The candidate is phase-internal cohesion: Items inside the same completed phase are spelled out as a unit and share locality-domain status, leading to more shared computation than structural depth alone predicts and making them representationally closer in the model's hidden states (\cref{sec:background}).

\paragraph{Why \textsc{esubj-evb} specifically.}
Phase cohesion would in principle apply to any pair of items inside the same phase; its predictive force differs across our two probe pairs for a structural reason.
For \textsc{esubj-evb}, both items lie inside the embedded clause, and the matrix phases contain them identically in every condition, cancelling against the bare baseline; only the embedded structure between and around them varies.
In the bare condition they occupy a single minimal domain with no embedded phase separating them --- the baseline.
In the infinitival condition an embedded \textit{v}P intervenes (the verb inside it, the subject above it) with no embedded CP containing both, so they straddle a phase edge and we observe $\binf > 0$.
In the finite condition the same embedded \textit{v}P intervenes, but an embedded CP now contains both; the shared CP tightens the pair more than the \textit{v}P loosens it, and we observe $\bfin < 0$ against the bare baseline.
For \textsc{wh-esubj}, the two items span clause boundaries and are never co-internal to any embedded phase, so phase cohesion has no effect.

We treat phase cohesion as one possible account.
The strong empirical claim is the sign asymmetry itself ($\bfin < 0$ and $\binf > 0$ in all 13 models, each at a majority of layers) which no \gls{ud}-, linear-, or monotone-complexity-based account predicts.

\subsection{Opposite-Signed Causal Effects: The Matrix-Verb Interchange}
\label{sec:causal}

The two observational signatures (\cref{sec:observational,sec:esubj-evb}) are not causal.
They show phase structure is \emph{encoded}, not that it is \emph{used}.
A single activation patch tests both, and removes the one confound the observational within-clause result cannot.

\paragraph{The confound an observational contrast cannot exclude.}
The \textsc{esubj-evb} asymmetry compares probe distances across strings that differ in surface form: The infinitival inserts \textit{to} between the two probed words, lengthening the clause.
An observational contrast therefore cannot fully separate phase structure from clause length, the intervening token, or the subject--verb agreement that co-varies with finiteness.
The activation patch below leaves the \emph{target string unchanged} (it edits only the model's internal representation of the matrix verb), so the tokens whose probe distance we measure, their linear positions, and the embedded clause's (non-)agreement are identical before and after.
A non-zero effect thus cannot arise from any property of the target string itself (its length, token frequency, agreement, or surface form) since all are held fixed.

\paragraph{Design.}
The matrix verb is the most immediate surface c-commander of both the embedded subject and the embedded verb (\cref{fig:trees}).
For a matched item pair differing only in matrix-verb class, we run the model on the target sentence and replace the residual stream at the matrix-verb position with the corresponding representation from the source sentence, patching at the earliest layers where the injected feature can propagate furthest downstream.\footnote{Both responses are strongest when patched at layer~1 and decay as the patch layer approaches the measurement layer.}
We then read $\dbeta = \beta_{\text{patched}} - \beta_{\text{target}}$ on \emph{both} probe pairs, each measured in its own reference band (\cref{sec:reference-layer}).
The interchange is run in both directions --- {finite$\rightarrow$infinitival} (inject a finite matrix verb into an infinitival target) and {infinitival$\rightarrow$finite} --- and, as a control, within the same condition (finite$\leftrightarrow$finite, infinitival$\leftrightarrow$infinitival), which changes the matrix verb's lexical identity but not the clause it selects.

\paragraph{Predictions.}
If the probe pairs track phase structure, one manipulation should move them in \emph{opposite} directions.
Injecting a finite verb adds a phase boundary between \emph{wh} and the embedded subject (pushing \textsc{wh-esubj} \emph{up}) while placing the embedded subject and verb inside a shared embedded CP (pulling \textsc{esubj-evb} \emph{down}); the infinitival direction mirrors both signs.
A same-class swap also perturbs the representation, but with no consistent sign, whereas the between-class swap has the predicted opposite signs.
The target retains its own surface cues (\textit{to}, agreement) that compete with the injected signal, so a directional effect is strong evidence while a null is uninformative.

\begin{figure}[t]
  \centering
  \includegraphics[width=\linewidth]{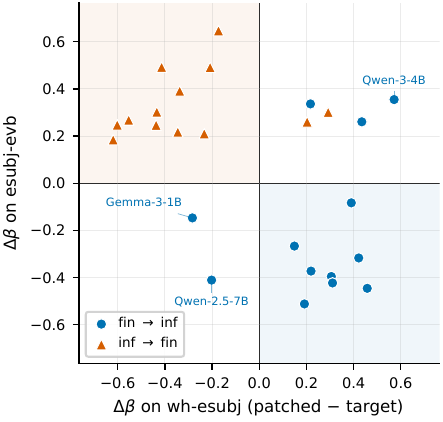}
  \caption{
    Opposite-signed causal effects from the matrix-verb interchange.
    Each point is one model in one interchange direction, plotted at its dominant in-band $\dbeta$ on \textsc{wh-esubj} (horizontal) and \textsc{esubj-evb} (vertical).
    Shaded quadrants mark the predicted regions ({finite$\rightarrow$infinitival} lower-right, \textsc{infinitival$\rightarrow$finite} upper-left); a point in the matching quadrant confirms the opposite-signed effect for that direction.
    The three labelled points are isolated exceptions (two cross-clause reversals and one within-clause reversal) discussed with the remaining cases in the text.
  }
  \label{fig:interchange-scatter}
\end{figure}

\paragraph{Results.}
\Cref{fig:interchange-scatter} plots the dominant in-band $\dbeta$ on the two probe pairs for each model and interchange direction; \cref{sec:interchange-forest} gives the underlying per-model values including controls.
A response counts as \emph{confirmed} when the sign of its dominant in-band $\dbeta$ is predicted; the effect is layer-localised (\cref{sec:reference-layer}), so the predicted sign need not hold at every band layer.
All four cells shift in the predicted direction in 8 of 13 models, and at least one interchange direction shows the opposite-signed shift in 11 of 13, with dominant $|\dbeta|$ typically $0.2$--$0.6$.
Pair by pair, the cross-clause \textsc{wh-esubj} effect is confirmed in both directions in 11 of 13 models and the within-clause \textsc{esubj-evb} effect in 10 of 13; the within-class controls carry no consistent sign and are, per pair, smaller than the between-class effect they control.
The failures are informative rather than noise.
In Gemma-3-1B and Qwen-2.5-7B, the cross-clause response reverses in-band in both directions, while the within-clause response behaves as predicted.
Gemma-3-27B, Qwen-3-1.7B, and Qwen-3-4B instead lose only the {finite$\rightarrow$infinitival} within-clause response; Qwen-3-4B shows a \emph{reversed} within-clause effect there ($\dbeta = +0.35$ where the prediction is negative), suggesting its within-clause asymmetry is more surface- or length-driven than phase-driven, precisely the alternative this experiment is built to detect.

\paragraph{Interpretation.}
On a monotone account, adding structure can only increase probe distances, so it cannot make one pair's distance fall while the other's rises.
The opposite-signed effect therefore requires pair-specific structure, of the kind phase-internal cohesion describes.
With the string held fixed, the same causal intervention which raises cross-clause separation also tightens phase-internal cohesion.
This is the strongest evidence we have that both signatures reflect one causally active phase structure rather than two surface correlations.

\section{Discussion}
\label{sec:discussion}

\paragraph{Structural encoding below the \gls{ud} surface.}
The \textsc{esubj-evb} sign asymmetry ($\bfin < 0$ and $\binf > 0$ in all 13 models, each at a majority of layers) is our clearest demonstration that \gls{llm} representations carry structural information the \gls{ud} probe target does not capture.
No \gls{ud}-distance, surface-linear, or monotone-complexity account predicts a sign reversal (\cref{sec:esubj-evb}); phase-internal cohesion does.
A same-verb control, holding the matrix verb fixed, further excludes a lexical-semantic account.
The finite tightening is also broad across the lexicon.
Broken down by embedded verb, $\bfin$ is negative for all 20 verbs in 11 of 13 models, and for at least 18 of 20 in the other two, so it is not carried by a few frequent subject-verb bigrams.
That it holds across four model families suggests distributional pretraining can induce representations aligned with formal-syntactic abstractions beyond the reach of annotation-based probing.

\paragraph{From correlation to one causal structure.}
The matrix-verb interchange (\cref{sec:causal}) turns two correlations into one causally active structure.
A single edit to the clause-selecting verb moves the cross-clause and within-clause pairs in opposite directions.
This is the two-way pattern phase-internal cohesion predicts, and no monotone account of depth, length, or complexity can produce it.
Because the edit leaves the target string fixed, it also closes the length-, agreement-, frequency-, and surface-cue loophole on the central asymmetry that an observational contrast must leave open.
The causal intervention shows that phase information is causally active \emph{among the model's representations}, not merely correlated with them; whether the model's \emph{behaviour} relies on this structure --- the sense at issue in \citet{agarwalMechanismsVsOutcomes2025} --- is a separate question we leave open.

\paragraph{Implications.}
Commonly adopted \gls{ud}-based parses provide merely a lower bound on syntactic encoding happening in \glspl{llm}. An invariant \gls{ud} tree distance does not mean a distinction is absent from the representation, only that it is invisible to a probe trained on that distance.
Substantively, that models trained only to predict tokens represent a phase-level organisation posited on independent linguistic grounds is a representational data point for the broader question of how far \gls{llm} internal structure aligns with generative syntax.

\section{Limitations}
\label{sec:limitations}

\paragraph{Language.}
All stimuli are in English.
The predictions follow from phase structure, which generalises cross-linguistically, but the three-way complement-size contrast exploits English-specific verb classes and case morphology.
Languages with overt morphological or word-order diagnostics for phase structure --- V2 languages, where finite verbs move to C, or ergative languages with distinct case--phase interactions --- would offer a more stringent test.

\paragraph{Model-level variability and scope.}
The causal responses are not universal: The within-clause effect fails or reverses in the {finite$\rightarrow$infinitival} direction for three models and the observational $\binf$ reverses in one, while the cross-clause response reverses in two further models (\cref{sec:observational,sec:causal}).
We have no principled account of why specific models depart.
The panel is 13 publicly available base models (1--27B) from four families; instruction-tuned, mixture-of-experts, and larger models are untested.

\section*{Acknowledgments}
This research was funded by the Defense Advanced Research Projects Agency (DARPA),
under contract W912CG23C0031.

\IfFileExists{writeup/references.trimmed.bib}{\bibliography{writeup/references.trimmed.bib}}{\bibliography{references.trimmed.bib}}

\clearpage
\appendix
\crefalias{section}{appendix}

\section{Full Model Panel}
\label{app:models}

\Cref{tab:models} lists all 13 models with their family, parameter count, and number of transformer blocks.
All models are used under licenses permitting academic research use: the Gemma Terms of Use (Gemma-3), the Meta Llama~3 Community License (Llama-3), and Apache~2.0 (Mistral-7B-v0.3, Qwen-2.5, Qwen-3).

\begin{table}[ht]
  \centering
  \small
  \caption{
    The 13-model panel, organised by family.
    ``Layers'' is the number of transformer blocks, excluding the embedding layer.
  }
  \label{tab:models}
  \begin{tblr}{
      colspec  = {llcr},
      width    = \linewidth,
      hline{1,Z} = {1pt},
      hline{2}   = {0.5pt},
      hline{6,8,9} = {0.25pt},
      rowsep   = 1.5pt,
    }
    Family & Model & Size & Layers \\
    Gemma-3  & Gemma-3-1B      & 1B   & 26 \\
    & Gemma-3-4B      & 4B   & 34 \\
    & Gemma-3-12B     & 12B  & 48 \\
    & Gemma-3-27B     & 27B  & 62 \\
    Llama-3  & Llama-3.2-1B    & 1B   & 16 \\
    & Llama-3.1-8B    & 8B   & 32 \\
    Mistral  & Mistral-7B-v0.3 & 7B   & 32 \\
    Qwen     & Qwen2.5-1.5B    & 1.5B & 28 \\
    & Qwen2.5-7B      & 7B   & 28 \\
    & Qwen3-1.7B      & 1.7B & 28 \\
    & Qwen3-4B        & 4B   & 36 \\
    & Qwen3-8B        & 8B   & 36 \\
    & Qwen3-14B       & 14B  & 40 \\
  \end{tblr}
\end{table}

\section{Syntactic Terminology}
\label{app:syntax-glossary}

This appendix glosses the syntactic terms used throughout the paper.

We use standard \textbf{X-bar} notation, in which a head X projects an intermediate level X\xbar{} that combines X with its complement, and a maximal phrase XP; the \textbf{specifier} (Spec,XP) is the phrase's structurally highest internal position.
The projections relevant here are the \textbf{CP} (complementiser phrase, headed by C, whose specifier Spec,CP is the landing site for wh-movement), the \textbf{TP} (tense phrase, whose specifier hosts the surface subject), the \textbf{\textit{v}P} (light-verb phrase, which introduces the external argument in Spec,\textit{v}P), the \textbf{VP} (lexical-verb phrase, containing the verb and its complement), and the \textbf{DP} (determiner phrase, the maximal nominal projection covering subjects, objects, and wh-phrases).
A node \textbf{c-commands} its sister and all its sister's descendants.

A \textbf{phase} is a cyclic spell-out domain; in \gls{mp} English the phase heads are the transitive light verb \textit{v} and the complementiser C, so \textit{v}P and CP are phases whereas TP and VP are not.
Once a phase is complete, the \textbf{Phase Impenetrability Condition} (\gls{pic}) makes its complement inaccessible to operations outside the phase, leaving only the phase head and its edge (the specifier) available.
Long-distance wh-movement must therefore be \textbf{successive-cyclic}, halting at each intervening phase edge --- Spec,\textit{v}P and Spec,CP --- on its way to the matrix Spec,CP \citep{chomskyDerivationPhase2001}.

Our stimuli turn on a small set of clause types.
A \textbf{small clause} is a minimal complement of only a VP and its subject, lacking TP and \textit{v}P, as in the bracketed portion of ``she saw [him eat the cake]''.
Under \textbf{exceptional case-marking} (\gls{ecm}) the matrix verb assigns accusative case to the overt subject of an infinitival complement --- \textit{him}, not \textit{he}, in ``she expected \textit{him} to leave''.
A \textbf{bridge verb} such as \textit{think} or \textit{say} permits wh-extraction from its finite CP complement, and the long-distance relation between the fronted wh-phrase and its extraction site is a \textbf{filler--gap dependency}.

\section{Detailed Phrase Structures}
\label{app:full-trees}

\newcommand{\FullTreeFinite}{%
  \begin{forest}
    for tree={s sep=2.5pt, inner sep=0.5pt, l=11pt, l sep=3pt, font=\scriptsize}
    [CP
      [DP [what$_i$]]
      [C\xbar
        [C [did]]
        [TP
          [DP [she]]
          [T\xbar
            [T [\sout{did}]]
            [\textbf{\textit{v}P}
              [DP [\sout{she}]]
              [\textit{v}\xbar
                [\textit{v} [think]]
                [VP
                  [V [\sout{think}]]
                  [\textbf{CP}, for tree={text=cFin}
                    [DP [\sout{what$_i$}]]
                    [C\xbar
                      [C [$\emptyset$]]
                      [TP
                        [DP [he]]
                        [T\xbar
                          [T [\sout{ate}]]
                          [\textbf{\textit{v}P}
                            [DP [\sout{he}]]
                            [\textit{v}\xbar
                              [\textit{v} [ate]]
                              [VP
                                [V [\sout{ate}]]
                                [DP [\sout{what$_i$}]]
                              ]
                            ]
                          ]
                        ]
                      ]
                    ]
                  ]
                ]
              ]
            ]
          ]
        ]
      ]
    ]
\end{forest}}

\begin{figure*}[t]
  \captionsetup[subfigure]{justification=raggedleft, singlelinecheck=false}
  \begin{subfigure}[t]{0.27\linewidth}
    \centering
    \begin{forest}
      for tree={s sep=3pt, inner sep=0.5pt, l=12pt, l sep=3pt, font=\scriptsize}
      [CP
        [DP [what$_i$]]
        [C\xbar
          [C [did]]
          [TP
            [DP [she]]
            [T\xbar
              [T [\sout{did}]]
              [\textbf{\textit{v}P}
                [DP [\sout{she}]]
                [\textit{v}\xbar
                  [\textit{v} [see]]
                  [VP
                    [V [\sout{see}]]
                    [VP, for tree={text=cBare}
                      [DP [him]]
                      [V\xbar
                        [V [eat]]
                        [DP [\sout{what$_i$}]]
                      ]
                    ]
                  ]
                ]
              ]
            ]
          ]
        ]
      ]
    \end{forest}%
    \makebox[0pt][l]{\vphantom{\FullTreeFinite}}%
    \caption{Bare}
  \end{subfigure}\kern-1.5em
  \begin{subfigure}[t]{0.36\linewidth}
    \centering
    \begin{forest}
      for tree={s sep=3pt, inner sep=0.5pt, l=12pt, l sep=3pt, font=\scriptsize}
      [CP
        [DP [what$_i$]]
        [C\xbar
          [C [did]]
          [TP
            [DP [she]]
            [T\xbar
              [T [\sout{did}]]
              [\textbf{\textit{v}P}
                [DP [\sout{she}]]
                [\textit{v}\xbar
                  [\textit{v} [expect]]
                  [VP
                    [V [\sout{expect}]]
                    [TP, for tree={text=cInf}
                      [DP [him]]
                      [T\xbar
                        [T [to]]
                        [\textbf{\textit{v}P}
                          [DP [\sout{him}]]
                          [\textit{v}\xbar
                            [\textit{v} [eat]]
                            [VP
                              [V [\sout{eat}]]
                              [DP [\sout{what$_i$}]]
                            ]
                          ]
                        ]
                      ]
                    ]
                  ]
                ]
              ]
            ]
          ]
        ]
      ]
    \end{forest}%
    \makebox[0pt][l]{\vphantom{\FullTreeFinite}}%
    \caption{Infinitival}
  \end{subfigure}\kern-1.5em
  \begin{subfigure}[t]{0.37\linewidth}
    \centering
    \FullTreeFinite
    \caption{Finite}
  \end{subfigure}
  \caption{
    Phrase structures with intermediate projections for the three conditions, following copy theory \citep{chomskyMinimalistInquiriesFramework2000}.
    Struck-through nodes are lower (unpronounced) copies: the auxiliary \textit{did} (C$\leftarrow$T), subjects (Spec,TP$\leftarrow$Spec,\textit{v}P), and verbs (v$\leftarrow$V).
    The wh-element (\textit{what}$_i$) is shown at its base position, at embedded Spec,CP (finite only), and at matrix Spec,CP; the intermediate copy at Spec,\textit{v}P of the embedded clause is omitted for clarity.
    The embedded clause is colour-coded by condition; phases crossed by the wh-element are in \textbf{bold}.
  }
  \label{fig:full-trees}
\end{figure*}

\Cref{fig:full-trees} gives phrase structures with intermediate projections for all three conditions, showing the \textit{v}P--VP decomposition under little-\textit{v} and copy-theoretic lower copies for subject raising and head movement; intermediate wh-movement copies are omitted for clarity.

\section{Stimulus Lexicon and Verification}
\label{sec:stimulus-appendix}

\begin{figure*}[t]
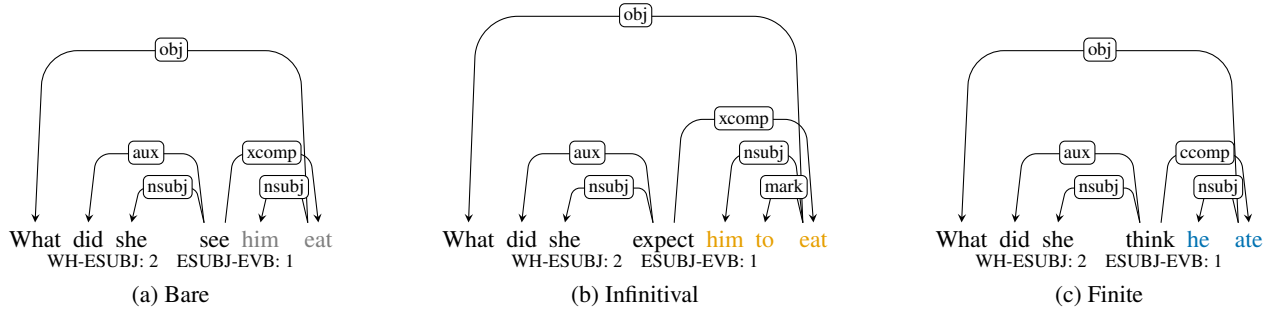

  \centering
  \begin{subfigure}[t]{0.31\linewidth}
    \centering
    \begin{dependency}
      \begin{deptext}[font=\footnotesize]
        What \& did \& she \&[1.5em] see \& \textcbare{him} \&[0.5em] \textcbare{eat} \\
      \end{deptext}
      \depedge{6}{1}{obj}
      \depedge{4}{2}{aux}
      \depedge{4}{3}{nsubj}
      \depedge{4}{6}{xcomp}
      \depedge{6}{5}{nsubj}
    \end{dependency}\\[-1em]
    {\scriptsize \MakeUppercase{wh-esubj}: 2\quad \MakeUppercase{esubj-evb}: 1}
    \caption{Bare}
  \end{subfigure}\hfill
  \begin{subfigure}[t]{0.35\linewidth}
    \centering
    \begin{dependency}
      \begin{deptext}[font=\footnotesize]
        What \& did \& she \&[1.5em] expect \& \textcinf{him} \& \textcinf{to} \&[0.5em] \textcinf{eat} \\
      \end{deptext}
      \depedge{7}{1}{obj}
      \depedge{4}{2}{aux}
      \depedge{4}{3}{nsubj}
      \depedge{4}{7}{xcomp}
      \depedge{7}{5}{nsubj}
      \depedge{7}{6}{mark}
    \end{dependency}\\[-1em]
    {\scriptsize \MakeUppercase{wh-esubj}: 2\quad \MakeUppercase{esubj-evb}: 1}
    \caption{Infinitival}
  \end{subfigure}\hfill
  \begin{subfigure}[t]{0.31\linewidth}
    \centering
    \begin{dependency}
      \begin{deptext}[font=\footnotesize]
        What \& did \& she \&[1.5em] think \& \textcfin{he} \&[0.5em] \textcfin{ate} \\
      \end{deptext}
      \depedge{6}{1}{obj}
      \depedge{4}{2}{aux}
      \depedge{4}{3}{nsubj}
      \depedge{4}{6}{ccomp}
      \depedge{6}{5}{nsubj}
    \end{dependency}\\[-1em]
    {\scriptsize \MakeUppercase{wh-esubj}: 2\quad \MakeUppercase{esubj-evb}: 1}
    \caption{Finite}
  \end{subfigure}
  \caption{
    Intended \gls{ud} parses of the three conditions, with embedded-clause words colour-coded by condition (matching \cref{fig:trees}).
    The design requires only that each probe pair's distance be constant across conditions --- two edges for \textsc{wh-esubj} ($\textit{what} \leftrightarrow \textit{him/he}$) and one edge for \textsc{esubj-evb} ($\textit{him/he} \leftrightarrow \textit{eat/ate}$) in this parse --- so any non-zero $\beta$ reflects structure beyond the \gls{ud} distance.
    An automatic parser realises this to within one edge for both pairs (\cref{tab:verify}).
  }
  \label{fig:ud-trees}
\end{figure*}

\paragraph{Lexicon.}
The combinatorial lexicon of \cref{sec:stimuli} has the following members.
Matrix subjects: \emph{she}, \emph{he}, \emph{they}, \emph{I}, \emph{the teacher}, \emph{the manager}, \emph{the detective}.
Embedded subjects, as matched accusative/nominative pairs: \emph{him}/\emph{he}, \emph{her}/\emph{she}, \emph{us}/\emph{we}, \emph{them}/\emph{they}, and \emph{the student}, \emph{the assistant}, \emph{the suspect} (identical in both cases).
Bare small-clause verbs: \emph{see} and \emph{watch} (visual perception), \emph{make} and \emph{let} (causative); \emph{hear} and \emph{have} are excluded (an auditory selectional restriction and an auxiliary clash, respectively).
Infinitival verbs: \emph{expect} (ECM), and \emph{want}, \emph{allow}, \emph{need} (object control).
Finite bridge verbs: \emph{think}, \emph{believe}, \emph{claim}, \emph{say}, \emph{know}, \emph{suppose}, \emph{report}.
Embedded transitive verbs, each taking an inanimate-compatible object, with a mix of regular and irregular past forms: \emph{eat}, \emph{buy}, \emph{read}, \emph{write}, \emph{build}, \emph{fix}, \emph{find}, \emph{clean}, \emph{cook}, \emph{sell}, \emph{carry}, \emph{paint}, \emph{break}, \emph{steal}, \emph{bring}, \emph{open}, \emph{send}, \emph{drop}, \emph{use}, \emph{take}.
Two constraints prune the Cartesian product: The matrix subject must differ from the embedded subject, and the three matrix verbs within an item must be distinct.
The result is a pool of roughly $128{,}000$ valid items, from which a fixed seed-controlled shuffle takes the first $1{,}000$.
An earlier version of the lexicon also listed \emph{convince} and \emph{require} as infinitival verbs; a scale check flagged both because spaCy attaches their post-verbal noun phrase as a direct object rather than an embedded subject, placing the probed pair two edges off the design in $37$ items, so they were removed and \emph{need} was substituted.

\paragraph{Distances and verification.}
The stimuli are templated so that each probed-pair distance is constant across conditions (\cref{fig:ud-trees}), verified per item on the automatic parse (\cref{tab:verify}).
To confirm this automatically, we parse all $3{,}000$ stimuli with spaCy \texttt{en\_core\_web\_trf}, compute the \gls{ud} tree distance from that parse for each probed pair in each condition, and record every item's \emph{spread}: the largest minus the smallest of its three condition distances.
An item passes when its spread is at most one edge.
\Cref{tab:verify} reports the fraction of items that are fully distance-invariant (spread $0$), for the main experiment and the same-verb control (\cref{sec:b1-appendix}).
All $3{,}000$ stimuli parse in both experiments, and no item exceeds the one-edge tolerance.

\begin{table}[t]
  \centering
  \small
  \caption{
    Stimulus verification: the percentage of the $1{,}000$ items whose automatic \gls{ud} distance is invariant across all three conditions (spread $0$), by probed pair, for the main experiment and the same-verb control (\cref{sec:b1-appendix}).
    Both parse all $3{,}000$ stimuli with spaCy \texttt{en\_core\_web\_trf}; every remaining item differs by a single edge, and none exceeds the one-edge tolerance.
  }
  \label{tab:verify}
  \begin{tblr}{
      colspec  = {l cc},
      hline{1,Z} = {1pt},
      hline{2}   = {0.5pt},
      rowsep   = 1.5pt,
    }
    Probed pair & Main & Control \\
    \textsc{wh-esubj}  & $79.4$ & $67.5$ \\
    \textsc{esubj-evb} & $98.8$ & $99.7$ \\
  \end{tblr}
\end{table}

\section{Probe Training Quality}
\label{app:probe-quality}

\begin{figure*}[t]
  \centering
  \includegraphics[width=\linewidth]{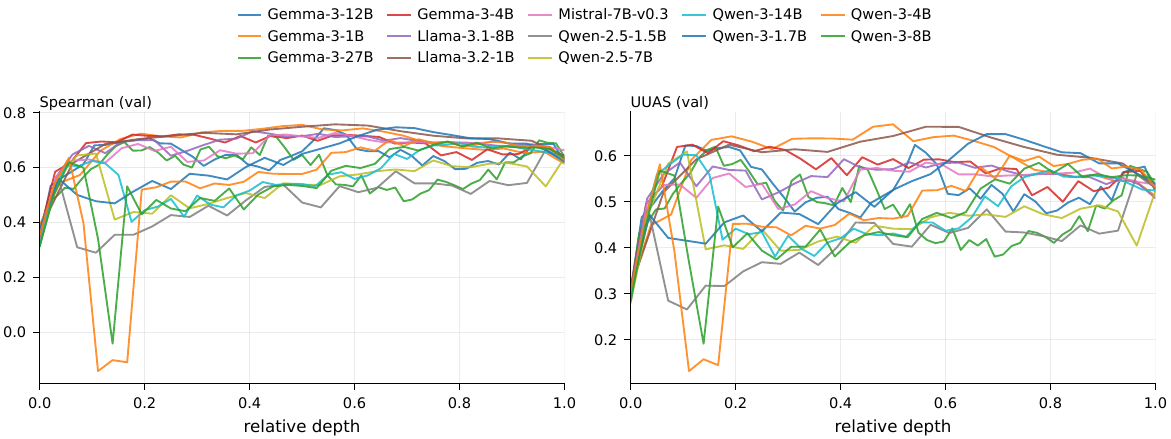}
  \caption{
    Per-layer probe quality on the \gls{udewt} validation set, one line per model.
    Left: distance Spearman correlation between predicted and gold undirected tree distances.
    Right: \gls{uuas}.
    All 13 probes follow the standard profile --- a rise from the embedding layer to a mid-network plateau, then a decline near the output.
    The horizontal axis is relative depth (layer index divided by block count) so models of differing depth are comparable.
  }
  \label{fig:probe-quality}
\end{figure*}

All 13 probes show the standard structural-probe quality profile on the \gls{udewt} validation set (\cref{fig:probe-quality}) --- a rise from the embedding layer to a mid-network plateau, then a moderate decline near the output --- with a peak distance Spearman correlation (predicted vs.\ gold tree distances) of $0.65$--$0.76$ and \gls{uuas} of $0.56$--$0.67$ across models, indicating substantial recovery of dependency structure.
Our analysis does not otherwise hinge on absolute probe quality --- the effect sizes $\beta$ are within-probe contrasts between conditions on the same probe, so a probe's overall recovery level bounds neither their sign nor their significance.

\section{Per-Model Observational Effects}
\label{sec:obs-forest}

\Cref{tab:obs-forest} gives the per-model within-clause \textsc{esubj-evb} effect behind \cref{fig:phase-forest}, the numeric basis for the sign asymmetry.
In every model the finite contrast is significantly negative and the infinitival contrast significantly positive at a majority of layers, so $\bfin < 0$ and $\binf > 0$ each hold panel-wide.

\begin{table}[t]
  \centering
  \small
  \caption{
    Per-model within-clause \textsc{esubj-evb} effect (\cref{sec:esubj-evb}).
    $\bfin$ and $\binf$ are the median contrast (finite\,$-$\,bare, infinitival\,$-$\,bare) across the layers that are \gls{fdr}-significant for that model, pair, and contrast --- the quantity plotted in \cref{fig:phase-forest}.
    The last two columns count the transformer blocks at which the contrast is \gls{fdr}-significant with the predicted sign ($\bfin < 0$; $\binf > 0$), out of the model's total; a majority in every model.
  }
  \label{tab:obs-forest}
  \begin{tblr}{
      colspec  = {l rr cc},
      hline{1,Z} = {1pt},
      hline{2}   = {0.5pt},
      hline{6,8,9} = {0.25pt},
      rowsep   = 1.5pt,
    }
    Model & $\bfin$ & $\binf$ & $\bfin<0$ & $\binf>0$ \\
    Gemma-3-1B      & $-0.16$ & $+0.15$ & 17/26 & 18/26 \\
    Gemma-3-4B      & $-0.15$ & $+0.24$ & 25/34 & 33/34 \\
    Gemma-3-12B     & $-0.21$ & $+0.25$ & 38/48 & 43/48 \\
    Gemma-3-27B     & $-0.18$ & $+0.16$ & 52/62 & 52/62 \\
    Llama-3.2-1B    & $-0.16$ & $+0.21$ & 14/16 & 16/16 \\
    Llama-3.1-8B    & $-0.14$ & $+0.28$ & 26/32 & 32/32 \\
    Mistral-7B-v0.3 & $-0.16$ & $+0.31$ & 26/32 & 26/32 \\
    Qwen2.5-1.5B    & $-0.10$ & $+0.17$ & 18/28 & 20/28 \\
    Qwen2.5-7B      & $-0.17$ & $+0.22$ & 23/28 & 23/28 \\
    Qwen3-1.7B      & $-0.18$ & $+0.23$ & 20/28 & 24/28 \\
    Qwen3-4B        & $-0.18$ & $+0.21$ & 32/36 & 32/36 \\
    Qwen3-8B        & $-0.19$ & $+0.10$ & 30/36 & 28/36 \\
    Qwen3-14B       & $-0.29$ & $+0.08$ & 38/40 & 22/40 \\
  \end{tblr}
\end{table}

\section{ECM versus Object-Control}
\label{app:ecm-control}

The infinitival class combines one \gls{ecm} verb (\emph{expect}) and three object-control verbs (\emph{want}, \emph{allow}, \emph{need}).
Splitting the within-clause infinitival contrast $\binf$ by subclass shows the two behave alike.
$\binf$ is positive in all 13 models for each subclass, of comparable magnitude (\cref{tab:ecm-control}), which justifies pooling the infinitival verbs into a single class in the main text.

\begin{table}[t]
  \centering
  \small
  \caption{
    Within-clause \textsc{esubj-evb} infinitival contrast $\binf$ (infinitival\,$-$\,bare), median across transformer blocks, split by the infinitival matrix verb's subclass.
    \gls{ecm} is \emph{expect}; object control is \emph{want}, \emph{allow}, and \emph{need}.
    For the object-control verbs the probed \textsc{esubj} token is the matrix object controlling a null PRO subject, not a syntactic embedded subject; the comparable, uniformly positive $\binf$ shows the two subclasses behave alike.
  }
  \label{tab:ecm-control}
  \begin{tblr}{
      colspec  = {l rr},
      hline{1,Z} = {1pt},
      hline{2}   = {0.5pt},
      hline{6,8,9} = {0.25pt},
      rowsep   = 1.5pt,
    }
    Model & $\binf$ (ECM) & $\binf$ (control) \\
    Gemma-3-1B      & $+0.09$ & $+0.14$ \\
    Gemma-3-4B      & $+0.19$ & $+0.23$ \\
    Gemma-3-12B     & $+0.28$ & $+0.23$ \\
    Gemma-3-27B     & $+0.19$ & $+0.11$ \\
    Llama-3.2-1B    & $+0.22$ & $+0.20$ \\
    Llama-3.1-8B    & $+0.27$ & $+0.27$ \\
    Mistral-7B-v0.3 & $+0.31$ & $+0.26$ \\
    Qwen2.5-1.5B    & $+0.17$ & $+0.13$ \\
    Qwen2.5-7B      & $+0.14$ & $+0.14$ \\
    Qwen3-1.7B      & $+0.16$ & $+0.24$ \\
    Qwen3-4B        & $+0.14$ & $+0.22$ \\
    Qwen3-8B        & $+0.07$ & $+0.09$ \\
    Qwen3-14B       & $+0.02$ & $+0.07$ \\
  \end{tblr}
\end{table}

\section{Same-Verb Control (Lexically Matched)}
\label{sec:b1-appendix}

\Cref{sec:esubj-evb} reports a same-verb control that holds the matrix verb fixed at \textit{expect} across the finite and infinitival conditions, isolating complement structure from matrix-verb identity and lexical semantics.
The stimuli reuse the main generator with the matrix verb locked to \textit{expect} in both the finite condition (\emph{What did she expect he ate?}) and the infinitival condition (\emph{What did she expect him to eat?}); the bare condition keeps a small-clause verb (\textit{see}, \textit{watch}, \textit{make}, \textit{let}) as the baseline, since \textit{expect} takes no bare complement.
We apply the trained probes to these stimuli without retraining.
Because the bare condition cannot share the matrix verb, only the finite--infinitival difference is strictly verb-matched, so we report the verb-matched contrast $\binf - \bfin$, which cancels the bare baseline.
\gls{ud}-distance invariance for the experimental pairs was verified on all 3{,}000 stimuli under the same criterion as the main stimuli (\cref{sec:stimuli}).

\Cref{tab:b1-forest} gives per-model estimates.
The verb-matched ordering $\binf - \bfin > 0$ is \gls{fdr}-significant with the predicted sign at a majority of layers in all 13 models, with median $\binf - \bfin$ between $0.19$ and $0.39$.
Against the cross-verb bare baseline $\binf$ is positive in every model, while $\bfin$ has a model-dependent sign --- \textit{expect} is a less prototypical finite-complement verb than the bridge verbs of the main experiment, so its finite arm sits closer to the baseline --- and the verb-matched difference, which combines both, is positive throughout.
This control isolates matrix-verb semantics; it does not address clause length, which the causal interchange (\cref{sec:causal}) removes separately.

\begin{table}[t]
  \centering
  \small
  \caption{
    Same-verb control (matrix verb fixed at \textit{expect}), \textsc{esubj-evb} pair.
    ``Layers'' is the number of transformer blocks at which the verb-matched contrast $\binf - \bfin > 0$ is \gls{fdr}-significant with the predicted sign, out of the model's total; a majority in every model.
    Medians are over \gls{fdr}-significant layers.
  }
  \label{tab:b1-forest}
  \begin{tblr}{
      colspec  = {lrrrr},
      row{1} = {valign = m},
      width    = \linewidth,
      hline{1,Z} = {1pt},
      hline{2}   = {0.5pt},
      hline{6,8,9} = {0.25pt},
      rowsep   = 1.5pt,
    }
    Model & Layers & {med.\\$\bfin$} & {med.\\$\binf$} & {med.\\$\binf-\bfin$} \\
    Gemma-3-1B      & 20/26 & $-0.08$ & $+0.09$ & $+0.21$ \\
    Gemma-3-4B      & 29/34 & $+0.03$ & $+0.20$ & $+0.22$ \\
    Gemma-3-12B     & 37/48 & $+0.10$ & $+0.28$ & $+0.25$ \\
    Gemma-3-27B     & 41/62 & $-0.02$ & $+0.20$ & $+0.20$ \\
    Llama-3.2-1B    & 16/16 & $-0.17$ & $+0.22$ & $+0.39$ \\
    Llama-3.1-8B    & 28/32 & $+0.10$ & $+0.26$ & $+0.19$ \\
    Mistral-7B-v0.3 & 28/32 & $+0.03$ & $+0.33$ & $+0.33$ \\
    Qwen2.5-1.5B    & 21/28 & $-0.01$ & $+0.18$ & $+0.22$ \\
    Qwen2.5-7B      & 24/28 & $-0.07$ & $+0.15$ & $+0.28$ \\
    Qwen3-1.7B      & 25/28 & $-0.11$ & $+0.13$ & $+0.26$ \\
    Qwen3-4B        & 34/36 & $-0.18$ & $+0.13$ & $+0.29$ \\
    Qwen3-8B        & 35/36 & $-0.23$ & $+0.08$ & $+0.27$ \\
    Qwen3-14B       & 39/40 & $-0.30$ & $+0.01$ & $+0.29$ \\
  \end{tblr}
\end{table}

\section{Per-Model Interchange Effects}
\label{sec:interchange-forest}

\begin{figure*}[t]
  \centering
  \includegraphics[width=\linewidth]{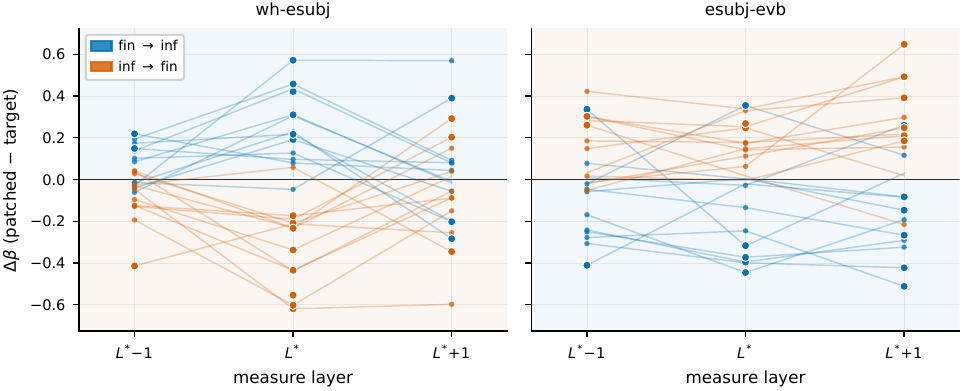}
  \caption{
    Within-band robustness of the interchange effect (\cref{sec:causal}).
    For each probe pair, $\dbeta$ (patched $-$ target) at the reference layer $L^{*}$ and its neighbours $L^{*}\pm1$: one line per model per interchange direction, with a filled dot at each significant band layer and a larger marker at the dominant (largest-magnitude significant) layer.
    The predicted half-plane for each direction is lightly shaded.
    The effect concentrates near $L^{*}$ rather than holding uniformly across the band, which is why we report the dominant in-band value (\cref{sec:reference-layer}); the two directions separate into opposite half-planes, and the few lines crossing into the wrong half-plane are the reversals flagged in \cref{tab:interchange-forest}.
  }
  \label{fig:interchange-robustness}
\end{figure*}

\Cref{tab:interchange-forest} gives the per-model coordinates behind \cref{fig:interchange-scatter}: for each interchange direction and probe pair, the dominant in-band $\dbeta$ (\cref{sec:causal}).
The opposite-signed prediction is that \textsc{fin$\to$inf} raises \textsc{wh-esubj} and lowers \textsc{esubj-evb}, and \textsc{inf$\to$fin} does the reverse.
All four cells carry the predicted sign in 8 of 13 models, and at least one interchange direction moves both pairs in their predicted opposite directions in 11 of 13; the within-class controls, which swap the matrix verb for another of the same phase class, carry no consistent sign and are far smaller.
\Cref{fig:interchange-robustness} profiles these effects across the reference band, showing they concentrate near $L^{*}$ rather than holding across it.

\begin{table*}[t]
  \centering
  \small
  \caption{
    Per-model matrix-verb interchange effects (\cref{sec:causal}), the coordinates behind \cref{fig:interchange-scatter}.
    Each cell is the dominant in-band $\dbeta$ (patched $-$ target) for one interchange direction and probe pair; the predicted sign is shown in the header ($+$: the probe distance increases).
    All reported values are significant (clustered bootstrap CI over items excludes $0$).
    $\dagger$ marks a cell whose dominant in-band effect significantly reverses the predicted sign.
    ``within'' is the larger of the two within-class (same-phase-class) control effects, $\max|\dbeta|$.
    ``effect'' summarises the pattern: \emph{both} $=$ all four cells predicted-signed; \emph{one} $=$ at least one direction moves both pairs oppositely; --- $=$ neither.
  }
  \label{tab:interchange-forest}
  \begin{tblr}{
      colspec  = {l rr rr r l},
      width    = \linewidth,
      hline{1,Z} = {1pt},
      hline{3}   = {0.5pt},
      hline{7,9,10} = {0.25pt},
      rowsep   = 1.5pt,
    }
    & \SetCell[c=2]{c} fin$\to$inf & & \SetCell[c=2]{c} inf$\to$fin & & & \\
    Model & wh-esubj\,$(+)$ & esubj-evb\,$(-)$ & wh-esubj\,$(-)$ & esubj-evb\,$(+)$ & within & effect \\
    Gemma-3-1B      & $-0.28^{\dagger}$ & $-0.15$ & $+0.29^{\dagger}$ & $+0.30$ & $0.03$ & --- \\
    Gemma-3-4B      & $+0.31$ & $-0.40$ & $-0.44$ & $+0.25$ & $0.02$ & both \\
    Gemma-3-12B     & $+0.19$ & $-0.51$ & $-0.21$ & $+0.49$ & $0.02$ & both \\
    Gemma-3-27B     & $+0.43$ & $+0.26^{\dagger}$ & $-0.43$ & $+0.30$ & $0.03$ & one \\
    Llama-3.2-1B    & $+0.42$ & $-0.32$ & $-0.55$ & $+0.27$ & $0.08$ & both \\
    Llama-3.1-8B    & $+0.22$ & $-0.37$ & $-0.41$ & $+0.49$ & $0.07$ & both \\
    Mistral-7B-v0.3 & $+0.15$ & $-0.27$ & $-0.17$ & $+0.65$ & $0.07$ & both \\
    Qwen2.5-1.5B    & $+0.39$ & $-0.08$ & $-0.35$ & $+0.22$ & $0.17$ & both \\
    Qwen2.5-7B      & $-0.20^{\dagger}$ & $-0.41$ & $+0.20^{\dagger}$ & $+0.26$ & $0.07$ & --- \\
    Qwen3-1.7B      & $+0.22$ & $+0.34^{\dagger}$ & $-0.23$ & $+0.21$ & $0.16$ & one \\
    Qwen3-4B        & $+0.57$ & $+0.35^{\dagger}$ & $-0.62$ & $+0.19$ & $0.04$ & one \\
    Qwen3-8B        & $+0.46$ & $-0.45$ & $-0.60$ & $+0.25$ & $0.03$ & both \\
    Qwen3-14B       & $+0.31$ & $-0.42$ & $-0.34$ & $+0.39$ & $0.02$ & both \\
  \end{tblr}
\end{table*}

\end{document}